\pdfoutput=1

\documentclass[11pt]{article}

\usepackage[]{acl}

\usepackage{times}
\usepackage{latexsym}
\usepackage{graphicx}
\usepackage{booktabs} 
\usepackage{multirow} 
\usepackage{algorithm}
\usepackage{algorithmic}

\usepackage[T1]{fontenc}

\usepackage[utf8]{inputenc}

\usepackage{microtype}

\usepackage{xcolor}
\usepackage{tcolorbox}
\tcbuselibrary{most}
\usepackage{colortbl}

\definecolor{lightgray}{gray}{0.9}

\usepackage{inconsolata}

\newtcolorbox{llmprompt}[1][]{
  colback=gray!20, 
  colframe=gray!70,
  fonttitle=\bfseries\ttfamily,
  title=#1, 
  rounded corners, 
  fontupper=\small\ttfamily,
  before upper={\raggedright}
}

\usepackage{arydshln}

\title{Think Twice: Perspective-Taking Improves Large Language Models' Theory-of-Mind Capabilities}

\author{Alex Wilf, Sihyun Shawn Lee, Paul Pu Liang, Louis-Philippe Morency \\
        Carnegie Mellon University \\ 
        \texttt{awilf@cs.cmu.edu} \\
}

\begin{document}
\maketitle
\begin{abstract}
Human interactions are deeply rooted in the interplay of thoughts, beliefs, and desires made possible by Theory of Mind (ToM): our cognitive ability to understand the mental states of ourselves and others.
Although ToM may come naturally to us, emulating it presents a challenge to even the most advanced Large Language Models (LLMs). Recent improvements to LLMs' reasoning capabilities from simple yet effective prompting techniques such as Chain-of-Thought (CoT)~\citep{wei2022chain} have seen limited applicability to ToM~\citep{gandhiUnderstandingSocialReasoning2023}. In this paper, we turn to the prominent cognitive science theory ``Simulation Theory'' to bridge this gap. We introduce \textsc{SimToM}, a novel two-stage prompting framework inspired by Simulation Theory's notion of \textbf{perspective-taking}. To implement this idea on current ToM benchmarks, \textsc{SimToM} first filters context based on what the character in question knows before answering a question about their mental state. Our approach, which requires no additional training and minimal prompt-tuning, shows substantial improvement over existing methods, and our analysis reveals the importance of perspective-taking to Theory-of-Mind capabilities. Our findings suggest perspective-taking as a promising direction for future research into improving LLMs' ToM capabilities. Our code is \href{https://github.com/shawnsihyunlee/simulatedtom}{publicly available}.
\end{abstract}

\section{Introduction}


What did the group of friends feel as they gathered around the fire, exchanging stories and laughter and knowing glances? Underlying this seemingly commonplace setting is an intricate interplay of thoughts, beliefs, and desires weaving together the fabric of human interaction. This is the domain of Theory of Mind (ToM): the cognitive ability to attribute mental states to ourselves and others, and to understand that others have beliefs, desires, and intentions that may differ from our own ~\citep{premack1978does, wellman2001}. This often unconscious ability is foundational to human cognition ~\citep{carruthers_2009} and social interaction ~\citep{langley2022theory}, yet
it is a task that, despite its simplicity, seems to perplex even the most advanced Large Language Models (LLMs)~\citep{gandhiUnderstandingSocialReasoning2023,sapNeuralTheoryofMindLimits2022}. Recently, simple prompting strategies such as Chain-of-Thought (CoT)~\citep{wei2022chain} have gained popularity because they can substantially improve LLM reasoning capabilities on some tasks without additional training or prompt tuning across models. Yet simple solutions to ToM still elude us~\citep{gandhiUnderstandingSocialReasoning2023}. Are LLMs incapable of performing ToM reasoning? Or have we just not found the right way to prompt them yet?

\begin{figure}[t]
\centering
\includegraphics[width=1.0\columnwidth]{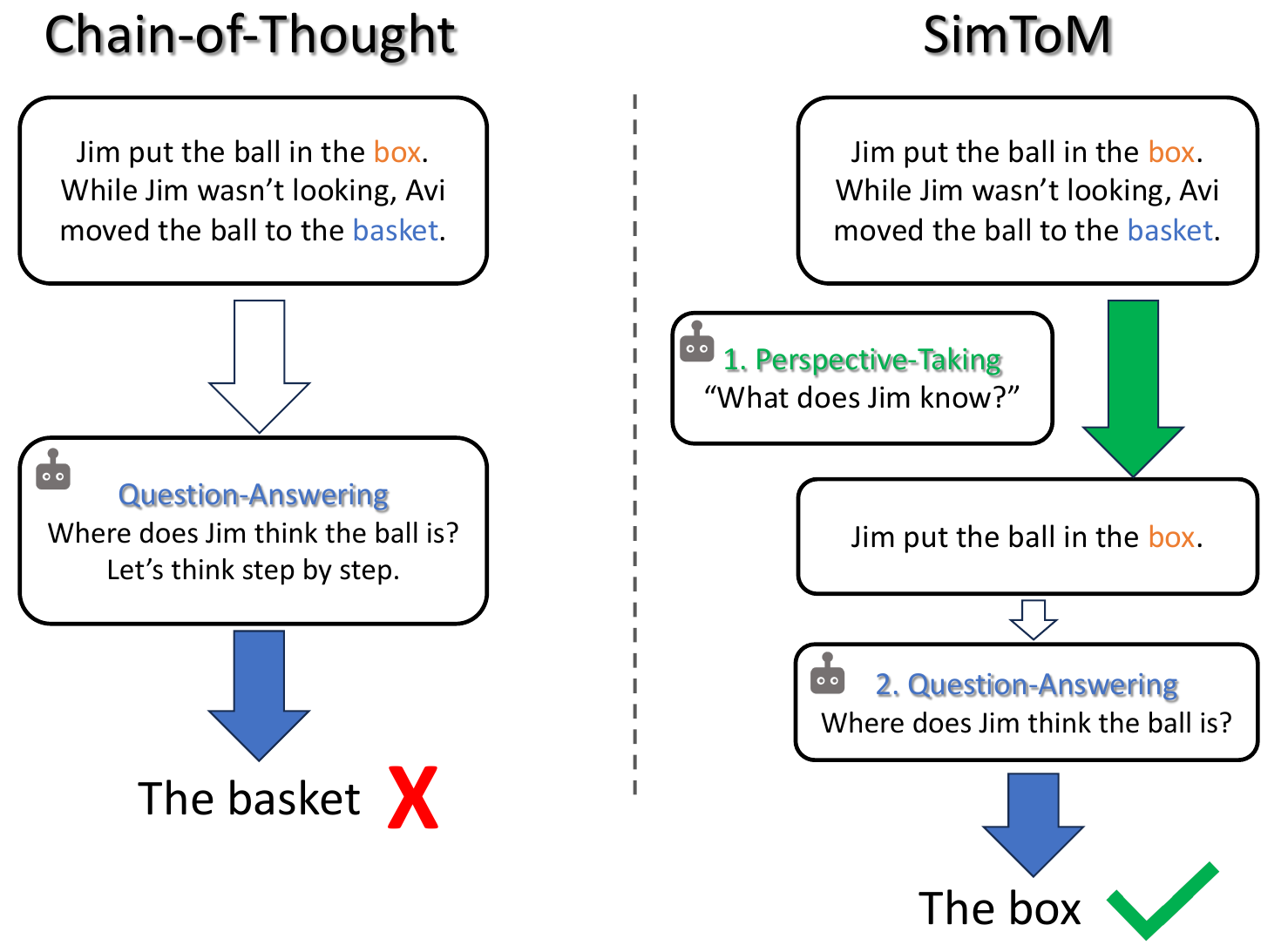}
\caption{Instead of performing Theory-of-Mind question-answering in a single inference pass, \textsc{SimToM} first prompts LLMs to perform \textit{perspective-taking}: filtering the context only to what the character in question \textit{knows}. Then, the LLM answers the question given this \textit{filtered} context. The example in this figure is representative of the core idea underlying current benchmarks used to gauge LLMs' ToM capabilities, called the Sally-Anne false-belief tests~\citep{BARONCOHEN198537}.}
\label{fig:intro}
\end{figure}




Although most current LLM probing strategies employ a single inference pass to answer ToM questions~\citep{gandhiUnderstandingSocialReasoning2023}, a prominent theory from cognitive science called ``Simulation Theory''~\citep{goldman2006} postulates that humans utilize a distinct step \textit{before} answering ToM questions called \textbf{perspective-taking} in which we ``step into the other person's shoes'', understanding their beliefs and goals before answering questions about their mental state ~\citep{barlassinaFolkPsychologyMental2017}. In the example in Figure~\ref{fig:intro}, understanding Jim's perspective amounts to understanding Jim's \textit{lack of knowledge} about a recent development (Avi moving the ball to the basket).


In this paper, we propose a simple two-stage prompting framework for LLMs inspired by Simulation Theory called \textsc{SimToM} that first implements perspective-taking, filtering the context only to what the person in question \textit{knows}, before answering Theory-of-Mind questions \textit{given that filtered context}.
Our approach seamlessly integrates with pre-trained LLMs, requiring no additional training and minimal prompt-tuning across models, while still demonstrating substantial performance improvements over off-the-shelf models using 0-shot MC and CoT probing.

We perform extensive analysis and ablations of our method and find that LLM's are surprisingly capable of perspective-taking when prompted and that improved perspective-taking capabilities are tied closely to \textit{further} improvements in ToM capabilities. These findings suggest that future research into Theory-of-Mind may find it useful to include \textsc{SimToM} as a simple yet effective baseline, and that this framework for thinking about ToM in LLMs may open new avenues for understanding and improving LLMs' abilities to simulate human-like ToM reasoning. Our code is  \href{https://github.com/shawnsihyunlee/simulatedtom}{publicly available}.

\section{Background}
\subsection{``Simulation'' Theory of Mind}
\label{subsec:sim_tom_psych}
``Simulation Theory'' (ST)~\citep{goldman2006} proposes an explanation for humans' ability to perform ToM that relies on a cognitive mechanism comprising two processes: \textbf{perspective-taking} (``putting yourself in their shoes''), followed by answering a ToM question from that perspective~\citep{hurley2008sim,goldman2008}. ST has strong philosophical~\citep{gordonAscentRoutinesPropositional2007,evansVarietiesReference1982,gordonFolkPsychologySimulation1986} and empirical supporte from decades of cognitive science research~\citep{galleseMirrorNeuronsSimulation1998,galleseUnifyingViewBasis2004, hurley2008sim}, though it is still an active area of debate (see Appendix~\ref{sec:appendix_st_tt} for a detailed discussion).

\paragraph{Perspective-Taking}
ST argues that perspective-taking, or placing oneself in another's position, is the initial step to simulating another's mental state. It involves simulating the beliefs and goals of the other individual. Crucial to this type of simulating are ``imagining believing'' what they believe~\citep{currieRecreativeMindsImagination2002,goldman2006}, or ``imagining desiring'' what they desire ~\citep{currieDesireImagination2002}. 

\paragraph{Question-Answering}
After perspective-taking, ST theorists argue that humans then answer a ToM question by observing and reasoning \textit{as if you were in their shoes} ~\citep{barlassinaFolkPsychologyMental2017,goldman2008}. Some theorists describe this as ``reuse'' of a ``cognitive mechanism'' ~\citep{hurley2008sim,craver2007} shared between humans.


\subsection{Are LLMs Capable of ToM?}
Supervised models can perform well on ToM tasks after finetuning , but~\citet{sclar-etal-2023-minding} show that they are brittle and overfit in ways that do not generalize to out-of-domain ToM tasks, suggesting that zero-shot methods may be more robust. As zero-shot methods and evaluation becoming increasingly common in NLP for this reason~\citep{zhao2023survey,sap-etal-2019-social}, we consider the unsupervised zero-shot setting for this work as well. 

Most modern LLMs struggle zero-shot on simple ToM tasks ~\citep{gandhiUnderstandingSocialReasoning2023, sapNeuralTheoryofMindLimits2022}. Some have claimed that recent ToM capabilities have emerged in large models~\citep{bubeck2023sparks,kosinski2023theory}, but others have argued that LLMs still fail on ``trivial'' alterations~\citep{ullman2023large} to existing datasets, suggesting limitations in current benchmark approaches or possible dataset leakage to closed-source models' training sets~\citep{shapira2023clever}.

Experimentally, current large models still lag behind human performance: for example, GPT-3.5-Turbo gets only 12.5\% on the ``action'' subset of false belief questions in BigTOM~\citep{gandhiUnderstandingSocialReasoning2023}. We find in Section~\ref{sec:results} that GPT-4 still lags behind human performance substantially on ToMI~\citep{le2019revisiting}, and although it performs well on BigTOM, this may be partly because GPT-4 itself was used to create the BigTOM dataset. From the literature and these results, it appears that LLMs do not yet reliably display zero-shot ToM capabilities~\citep{gandhiUnderstandingSocialReasoning2023}.

\section{Benchmarking Theory-of-Mind Capabilities}
\label{sec:datasets}
One well studied method for evaluating theory of mind capabilities is through the Sally Anne false-belief tests~\citep{BARONCOHEN198537}. In essence, one agent (Sally) knows something about the world, then they leave, and another agent (Anne) changes something about the world. For example: Sally puts a ball in the basket then leaves the room, after which Anne moves the ball to the box. 


We can then ask a few different types of questions, for example: ``Where does Sally believe the ball is?'' If Anne has moved the ball, Sally's belief will be incorrect – this type of question is called \textbf{false belief}, and has its counterpart in \textbf{true belief} questions, where Sally's belief about the world is correct. We can also ask about \textbf{actions} Sally would take as a result of those beliefs, for example: ``What will Sally do when she returns looking for the ball?''. And instead of asking about Sally directly, we could also ask about what \textbf{Anne thinks Sally thinks} – this is called a \textbf{second order} question, contrasted with the \textbf{first order} questions above.

To the best of our knowledge, there are two existing datasets that test these capabilities in the reading comprehension setting: \textbf{ToMI} and \textbf{BigTOM}.\footnote{Both datasets are available in the English language only.}

\subsection{ToMI}
\label{subsec:tomi}
ToMI~\citep{le2019revisiting} is a dataset of Sally-Anne stories, questions, and answer choices.\footnote{Made publicly available with the 
\href{https://github.com/facebookresearch/ToMi/blob/master/LICENSE}{CC License}.} For this paper, we use the updated version of ToMI from ~\citep{arodiTextualTimeTravel2021,sapNeuralTheoryofMindLimits2022} that has relabelled mislabelled second-order questions and disambiguated the location of containers after their reference (e.g., ``The ball is in the \textbf{basket}. The \textbf{basket} is in the front yard.''). 

\begin{figure*}[ht]
\centering
\includegraphics[width=2.0\columnwidth]{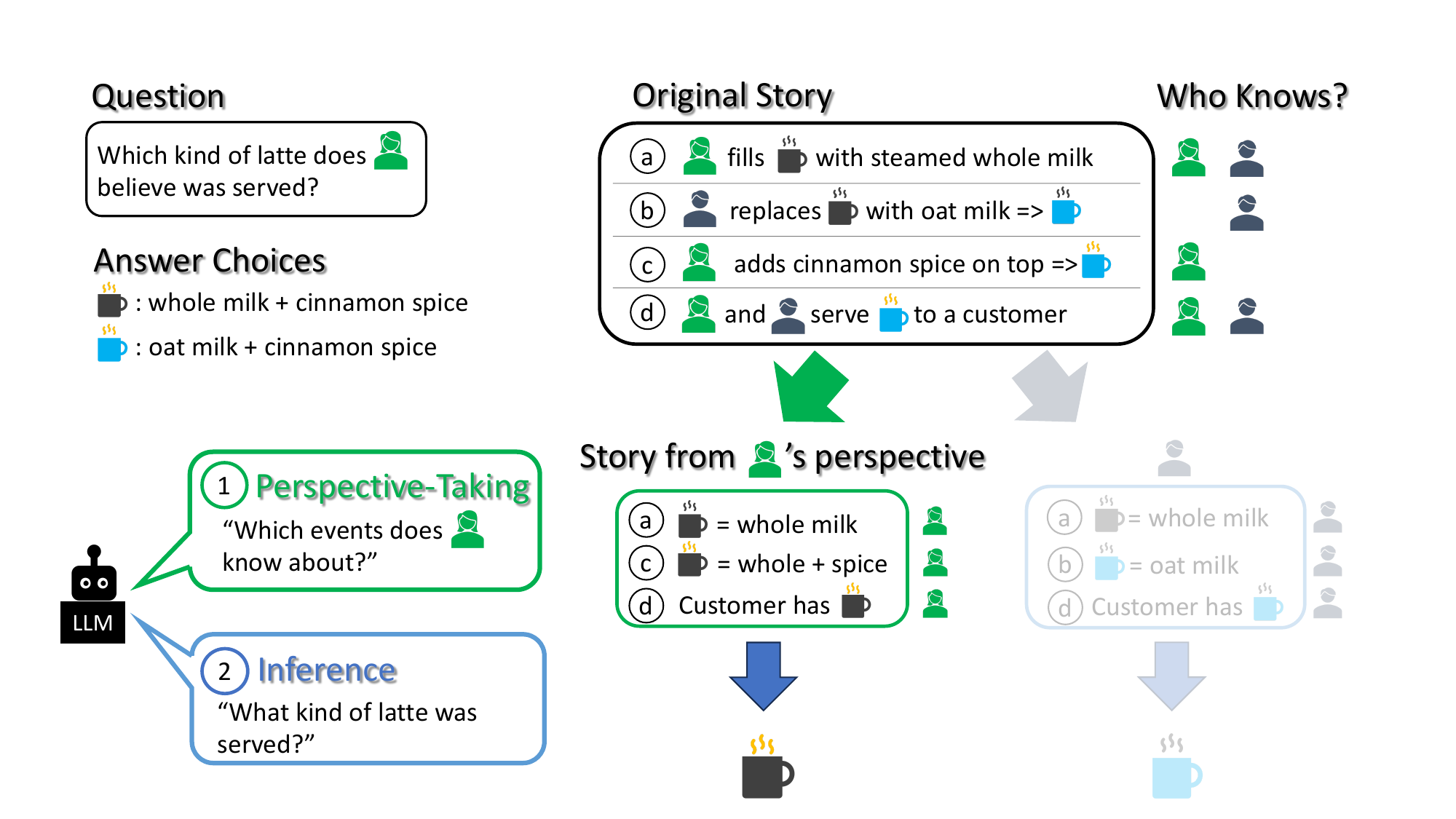}
\caption{An overview of \textsc{SimToM}, a two-stage prompting framework for enhancing zero-shot Theory-of-Mind capabilities in LLMs. The first step is \textbf{perspective-taking}, in which a model attempts to understand what the agent knows and wants. We then query the LLM to \textbf{infer} the answer to the question given this \textit{perspective}.
}
\label{fig:method}
\end{figure*}

\subsection{BigTOM}
\label{subsec:bigtom}
BigTOM~\citep{gandhiUnderstandingSocialReasoning2023} is also a Sally-Anne false belief-style ToM benchmark.\footnote{Made publicly available with the \href{https://github.com/cicl-stanford/procedural-evals-tom/blob/main/LICENSE}{MIT license}.} However, BigTOM evaluates ToM capabilities on a larger space of tasks than modification in object location and frames its stories in more natural language and social settings. BigTOM achieves this by building a causal template defining an agent's desire, percept, and initial belief, before generating a causal event that changes the environment and generating the resulting agent's belief or action. The authors of BigTOM create these templates, and generate the outputs using GPT-4. 

\section{\textsc{SimToM}: \textsc{Simulated} Theory of Mind}
\label{sec:methods}
\textsc{SimToM} is a simple two-stage prompting framework for that enhances zero-shot ToM capabilities in LLMs.

\subsection{Motivation}
We illustrate a motivating example in Figure~\ref{fig:method}.\footnote{The example we use is very similar to an actual question from BigTOM~\citep{gandhiUnderstandingSocialReasoning2023}, although with two false beliefs instead of one.} The story is as follows: the woman in green fills a cup with steamed whole milk, after which the woman \textit{does not see} the man in purple replace the whole milk in the cup with oat milk. The woman then adds cinnamon spice on top, which \textit{the man does not see}, then \textit{both} observe that the customer receives their drink. The question is ``Which kind of latte does the woman in green believe was served? Whole milk + cinnamon spice, or oat milk + cinnamon spice?'' The correct answer is whole milk + cinnamon spice, because the woman is not aware of the change the man made. 

0-shot CoT prompting will pass the whole story in as context and ask the LLM to reason through the answer:
\begin{llmprompt}[CoT Prompting]
\{story\} \\
\{question\} \\
\{answer choices\} \\
Answer the question based on the context.\\
Reason step by step before answering. \\
\end{llmprompt}

However, CoT will often output the \textit{true answer} – in this case, the type of latte the customer \textit{actually received}: oat milk + cinnamon spice. This amounts to a failure of perspective-taking: answering the question based on \textit{what she knows and what she does not know, regardless of whether it is correct or not}.


Motivated by this intuition and the literature on Simulation Theory, we hypothesize that LLMs' may be having difficulty with ToM reasoning because they are attempting to perform two tasks in a \textit{single inference pass}: perspective-taking and question-answering. To solve this, we break the ToM reasoning process into two inference passes:
\begin{enumerate}
    \item \textbf{Perspective-Taking}: understand what the woman knows
    \item \textbf{Question-Answering}: answer the question given \textit{what the woman knows} (\textit{not} the whole story)
\end{enumerate}



\subsection{Perspective-Taking}
~\citet{barlassinaFolkPsychologyMental2017} describe Perspective-Taking as ``switching roles'' to understand the other person's ``relevant beliefs and goals''. In \textsc{SimToM}, we implement this in a simple, concrete way: by asking models to first \textit{filter} the story to only the events that the character in question knows about.\footnote{Our implementation of \textsc{SimToM} requires the name of the character the question asks about – e.g., ``the woman in green''. We parse this during preprocessing, described in Section~\ref{sec:experiments}.}. To do this, we prompt an LLM as follows:

\begin{llmprompt}[\textsc{SimToM} Step \#1: Perspective-Taking]
The following is a sequence of events: \{story\} \\
Which events does \{character\_name\} know about?
\end{llmprompt}

\subsection{Question-Answering}
Question-Answering proceeds just as in baseline 0-shot or CoT, except that we replace the \textit{full story} with our \textit{modified} version resultig from Perspective-Taking. The story is modified so that information that the agent does not know about is \textit{hidden} from the LLM when it answers the question. In this example, the woman does not know that the man swapped the milk, so that information is removed during perspective-taking.

\begin{llmprompt}[\textsc{SimToM} Step \#2: Question-Answering]
\{story from character\_name's perspective\} \\
Answer the following question: \\
\{question\}
\end{llmprompt}

In this example, the story from the woman in green's perspective is: she fills a cup with steamed whole milk, adds cinnamon spice, and serves it to a customer. During Question-Answering, we would prompt the model \textit{only} with this perspective, then ask the same question of the model.

\begin{table*}[ht!]

\centering
\caption{\textsc{SimToM} results on BigTOM and ToMI across False Belief and All question types. We include the \textbf{absolute accuracy difference} between \textsc{SimToM} and the baselines (0-shot and 0-shot CoT) in parentheses.}
\label{tab:model_results}

\begin{tabular}{lcccc}
\toprule
& \multicolumn{2}{c}{\textbf{False Belief}} & \multicolumn{2}{c}{\textbf{All}} \\
\cmidrule(lr){2-3} \cmidrule(lr){4-5}
\textbf{Method} & BigTOM & ToMI & BigTOM & ToMI \\
\midrule

\multicolumn{5}{l}{\cellcolor{lightgray}0-Shot} \\

Llama2-7b-chat & 47.5 & 28.25 & 53.62 & 44.5 \\
Llama2-13b-chat & 41.25 & 39.25 & 51.38 & 51.0 \\
gpt-3.5-turbo & 41.0 & 67.25 & 66.38 & 68.6 \\
gpt-4 & 89.0 & 25.5 & 92.5 & 66.5 \\
[0.5em]

\multicolumn{5}{l}{\cellcolor{lightgray}0-shot CoT} \\
Llama2-7b-chat & 31.5 & 24.0 & 48.62 & 43.7 \\
Llama2-13b-chat & 52.25 & 16.5 & 56.0 & 45.0 \\
gpt-3.5-turbo & 56.25 & 34.0 & 75.88 & 64.1 \\
gpt-4 & 93.25 & 74.25 & 95.5 & 74.4 \\
[0.5em]

\multicolumn{5}{l}{\cellcolor{lightgray}\textsc{SimToM}} \\

Llama2-7b-chat & 70.5 (\textbf{+23.0, 39.0}) & 40.0 (\textbf{+11.8, 16.0}) & 57.25 (\textbf{+3.6, 8.6}) & 48.1 (\textbf{+3.6, 4.4}) \\
Llama2-13b-chat & 61.75 (\textbf{+20.5, 9.5}) & 35.5 (\textbf{-3.8, +19.0}) & 58.0 (\textbf{+6.6, 2.0}) & 61.1 (\textbf{+10.1, 16.1}) \\
gpt-3.5-turbo & 70.5 (\textbf{+29.5, 14.2}) & 81.0 (\textbf{+13.8, 47.0}) & 81.62 (\textbf{+15.2, 5.7}) & 72.8 (\textbf{+4.2, 8.7}) \\
gpt-4 & 92.0 (\textbf{+3.0, -1.2}) & 87.75 (\textbf{+62.2, 13.5}) & 95.0 (\textbf{+2.5, -0.5}) & 87.8 (\textbf{+21.3, 13.4}) \\

\bottomrule
\end{tabular}

\end{table*}

\section{Experimental Details}
\label{sec:experiments}
Our experiments are intended to investigate the effectiveness of our method by evaluate LLMs 0-shot on ToM benchmarks with and without our \textsc{SimToM} prompting framework. In this section, we detail how models are prompted and evaluated on the benchmarks described in Section~\ref{sec:datasets}: BigTOM and ToMI.

\subsection{Prompting}
We evaluate LLMs using MC-probing: we prompt models with a story, a question, and answer choices, and ask it to choose the correct answer choice given the question and story. Models can decline to answer (and the Llama models often do this), which can reduce their performance below 50\% random accuracy. We reproduce our 0-shot prompts exactly in Appendix~\ref{sec:appendix_baseline_prompts}. 

Our prompts for \textsc{SimToM} vary somewhat based on the structure of the dataset, and vary minimally between Llama and GPT models. We reproduce the exact prompts used in Appendix~\ref{sec:appendix_prompts}.  


\subsection{Evaluation}
Because false belief questions are the most challenging question category for modern models~\citep{gandhiUnderstandingSocialReasoning2023} and are at the core of the ``Sally-Anne False-Belief Tests'', we report our results averaged across all false belief question types and across All question types in our results. For BigTOM, this means averaging across ``Forward Action'' and ``Belief'' false belief questions; for ToMI, this includes averaging across first and second order question types.

We evaluate our approach on four state of the art language models: two open source – Llama2-7b and 13-b chat ~\citep{touvron2023llama} – and two closed source: GPT-3.5-Turbo and GPT-4~\citep{openai2023gpt4}. We query all models with temperature=0.0 for reproducibility. We run inference on the open source models on a single A100 GPU, and query the closed-source models using their API. The result for one model on one benchmark takes around three hours to run.

\paragraph{BigTOM}
In our experiments we consider the ``Forward Action'' and ``Forward Belief'' questions and not the ``Backward Belief'' so as to closely mimic the structure of the ToMI questions. BigTOM is balanced, with 200 questions per question type, so we do not randomly sample as we do for ToMI. BigTOM is a binary MC task, with random accuracy being 50\%. We parse the character name deterministically: it is the first word in each story.

\paragraph{ToMI}
To evaluate our methods on ToMI~\citep{le2019revisiting}, we randomly sample 100 samples from each of the ten question types to create a balanced dataset of 1000 samples. We parse the name of the character in question deterministically: because ToMI is created from templates, the character name is always the third word in the question.

The original ToMI dataset does not include the answer choices in the questions. We find that this artificially depresses baseline performance, as LLMs often output ambiguous answers. For example, an LLM outputted this answer to a ToMI question: ``Charlotte look for the melon in the \textbf{front yard}'', despite the only valid choices being either container A or B (both situated in the front yard). To rectify this, we parse the ToMI stories and include both possible answer choices (e.g., A and B) from the template used for story creation when querying the model. This process makes this setting a binary multiple-choice task similar to BigTOM. A random guessing baseline would have 50\% accuracy.

\begin{table*}[ht!]
\centering
\caption{Ablation Analysis of \textsc{SimToM}: we find that performing perspective-taking and question-answering in two prompts is necessary, as \textsc{SimToM}-Single does not lead to the same performance improvements over baselines. 
We also find that \textit{perspective-taking} is important, as \textsc{SimToM}-Multi implements a two-step prompting asking first for rationale.
By contrast, domain-specific perspective-taking prompting (\textsc{SimToM}-Domain) designed to simulate more advanced perspective-taking systems, leads to substantial performance gains, and oracle perspective-taking (\textsc{SimToM}-Oracle) from human-annotated perspectives can help today's LLM's nearly solve current ToM benchmarks. +/- values are absolute accuracy differences relative to \textsc{SimToM}, described in Section~\ref{sec:methods}.}
\label{tab:ablation_results}

\begin{tabular}{lcccc}
\toprule
& \multicolumn{2}{c}{\textbf{False Belief}} & \multicolumn{2}{c}{\textbf{All}} \\
\cmidrule(lr){2-3} \cmidrule(lr){4-5}
\textbf{Method} & \textbf{ToMI} & \textbf{BigTOM} & \textbf{ToMI} & \textbf{BigTOM} \\
\midrule
0-shot & 67.25 & 41.0 & 68.6 & 66.38 \\
0-shot CoT & 34.0 & 56.25 & 64.1 & 75.88 \\
\textsc{SimToM} & 81.0 & 70.5 & 72.8 & 81.62 \\
\hdashline
\textsc{SimToM}-Single & 58.75 (-22.25) & 50.75 (-19.75) & 67.5 (-5.3) & 54.75 (-26.87) \\
\textsc{SimToM}-Multi & 31.75 (-49.25) & 37.25 (-33.25) & 46.8 (-26.0) & 58.25 (-23.37) \\
\textsc{SimToM}-Domain & 85.5 (+4.5) & 90.5 (+20) & 79.3 (+6.5) & 91.5 (+9.88) \\
\textsc{SimToM}-Oracle & 96 (+15) & 96 (+25.5) & 82 (+9.2) & 98 (+16.38) \\
\bottomrule
\end{tabular}

\end{table*}

\section{Results and Discussion}
\label{sec:results}
We find that \textsc{SimToM} leads to substantial performance improvements over 0-shot MC and CoT prompting.
Our results in Table~\ref{tab:model_results} reflect these gains across both the BigTOM and ToMI benchmarks.

\paragraph{BigTOM Results}
On the BigTOM benchmark, \textsc{SimToM} provides substantial performance improvements across models. Notably, on the challenging false belief subset, \textsc{SimToM} provides \textbf{29.5\%} and \textbf{14.2\%} absolute accuracy improvements over the 0-shot and 0-shot CoT GPT-3.5-Turbo baselines, along with similarly strong results across other model types. \textsc{SimToM} led to a slight decrease with GPT-4, though this result may be partially confounded by the fact that BigTOM was generated using GPT-4. Across all question types as well, \textsc{SimToM} performed strongly compared with 0-shot CoT probing, leading to \textbf{8.6\%} and \textbf{5.7\%} absolute improvements on Llama2-7b-chat and GPT-3.5-Turbo for False Belief questions. We again saw comparable performance with the strongest model, GPT-4, which had saturated human performance 0-shot~\citep{gandhiUnderstandingSocialReasoning2023}.

\paragraph{ToMI Results}
\textsc{SimToM} also yields improvement over 0-shot baselines across LLMs on the ToMI benchmark. Compared to 0-shot CoT on false belief (FB) questions \textsc{SimToM} increased absolute accuracy by 16\%, 19\%, and a surprising \textbf{47\% absolute accuracy} for Llama2-7b-chat, Llama2-13b-chat, and GPT-3.5-Turbo. It is worth noting that 0-shot CoT did not perform reliably better than 0-shot, though our method performed better than both most of the time.
Across all question types, \textsc{SimToM} also provided performance increases as well, although more modest, as modern models find control conditions much less challenging (e.g., ``True Belief'' tests, which can be answered by understanding the true state of objects \textit{without} understanding mental states relating to them). These improvements were still substantial, with GPT-4 seeing a 13.4\% improvement and Llama-2-13b-chat improving by 16.1\% over 0-shot CoT probing. Additionally, we tested two additional prompting strategies: Self-Consistency CoT~\citep{wang2022self} and Tree-of-Thoughts~\citep{yao2023tree} on the GPT-3.5-Turbo model. SC-CoT performs similarly to CoT across question categories, underperforming \textsc{SimToM} significantly (33.50\% on FB compared to 81.00\%). Tree-of-Thoughts also underperforms \textsc{SimToM} substantially, perhaps because the original Tree-of-Thoughts voting evaluates creative writing generations instead of rationales. In general, we find that second-order questions are more difficult for most models than first-order questions; a detailed breakdown of these results across question types can be found in Appendix~\ref{sec:appendix_breakdown}.

\section{Analysis}
\label{sec:analysis}
We rigorously analyze \textsc{SimToM} with additional experiments and ablations intended to better understand our method explore the opportunities and it uncovers.

\subsection{Ablation Study: Single-Prompt \textsc{SimToM}}
\label{subsec:anal_sim_oneprompt}
Given \textsc{SimToM}'s strong performance, we are curious to determine whether our intuition from Section~\ref{sec:methods} is correct: that performing perspective-taking in a \textit{separate} prompt before question-answering is important for enhancing LLMs' ToM capabilities. 

To evaluate this, we implement a one-prompt ablation of \textsc{SimToM}, \textsc{SimToM}-Single, that combines perspective-taking and question-answering in a single prompt. A simplified version of our prompt is below, and our prompts are reproduced in full in Appendix~\ref{sec:appendix_single_prompts}. We run our ablation experiments using GPT-3.5-Turbo.
Our results are depicted in Table~\ref{tab:ablation_results}.

\begin{llmprompt}[\textsc{SimToM}-Single]
Your task is in two steps. \\
Step 1. output only the events that \{character\_name\} knows about. \\
Step 2. Imagine you are \{character\_name\}, then answer a question based only on the events \{character\_name\} knows about. \\
Story: \{story\} \\
Question: \{question\} \\
\end{llmprompt}

Interestingly, we find that performing perspective-taking before question-answering in a single prompt is \textbf{not nearly as effective as performing the same process with two prompts}. Qualitatively, we find that single-inference approaches (i.e., 0-shot, 0-shot CoT, and \textsc{SimToM}-Single) fail in similar ways: LLMs prompted in this way often answer the question with respect to the world state instead of with respect to the person's mental state. For example, in the example from Figure~\ref{fig:method}, LLMs would output ``oat milk + cinnamon spice''. This result supports our intuition from Section~\ref{sec:methods}, that LLMs may benefit from a \textit{separate} perspective-taking step when performing ToM reasoning tasks.

In addition to ablating our method, we were curious whether SimTOM's strong performance was a result of its multi-turn nature instead of the exact structure of its multistep prompting strategy. To evaluate this, we also attempted a multi-step version of CoT: in the first step asking the model to output its reasoning, thinking ``step by step'', then in the second asking it to answer a question given this reasoning. This strategy effectively replaces our ``perspective-taking'' with a ``reasoning'' step, holding the rest of the experiment constant. The performance of the 0-shot CoT model plummeted; reducing to 58.25\% on BigTOM total (a reduction of 17.63\% from 0-shot CoT and a reduction of 23.37 from \textsc{SimToM}) and to 46.8\% on ToMI (a reduction of (17.3\%). This indicates that the strength of SimTOM is not purely in its multi-step nature, but in the perspective-taking step specifically.


\subsection{Oracle Perspective-Taking}
\label{subsec:anal_oracle}
In this analysis, we investigate perspective-taking further by asking how much of the remaining performance gap \textit{can be improved with more effective perspective taking}. Put another way, can today's models answer ToM questions in ToMI and BigTOM given a well-filtered story? If so, perspective-taking may be justifiably described as a core challenge in future ToM research.



To evaluate this, we create a small dataset of ``oracle perspectives'': we ask four paid human annotators to output the ``perspective'' given the story and the name of the person, then use models to perform question-answering \textit{contingent on this perspective, a human-edited subset of the story}. We give annotators the same perspective-taking prompts as we give to GPT-3.5-Turbo (see Appendix~\ref{sec:appendix_prompts}), and additionally instruct them to output only a subset of the story, not add any additional information. Just as in Section~\ref{sec:methods}, they are not shown the question or answer choices during the process of perspective-taking. To construct this dataset, we sample from the benchmarks in a balanced manner: 100 questions from each question type across both benchmarks.


We find that with oracle perspective-taking, models can infer substantially better than with model-generated perspective-taking: improving on ToMI False Belief from 81.0\% to 96\% accuracy, and on BigTOM False Belief from 70.5\% to 96.0\% accuracy, close to solving these challenging false belief tasks. Performance lags behind on all ToMI questions, likely because of the different question types that do not specifically test this kind of simulation, including true belief, ``memory'' and ``reality''-style questions. Our analysis results are shown in Table~\ref{tab:ablation_results}, in the row \textsc{SimToM}-Oracle. From this we conclude that \textbf{human-level perspective-taking capabilities could enable today's LLM's to nearly solve current ToM benchmarks}.

\subsection{Domain-Specific Perspective-Taking Strategies}
\label{subsec:anal_domain_specific}
We next investigate to what extent current LLM's are capable of improving their perspective-taking capabilities.

To explore this question, we simulate how LLMs with advanced perspective-taking capabilities would perform on ToM by enhancing the \textsc{SimToM} with 3 examples of perspective-taking drawn from in-domain ToMI stories. 
The prompts follow the simplified form below. For the exact prompts, please refer to Appendix~\ref{sec:appendix_domain_specific_prompts}.

\begin{llmprompt}[\textsc{SimToM}-Domain]
Story:\\
1 Jackson is wearing the pajamas \\
2 Logan entered the dining room \\
... \\
Which events does Logan know about? \\
Logan knows about the following events: \\
2 Logan entered the dining room \\
...
\end{llmprompt}





Because we include examples of perspective-taking, we refer to this result as Domain-Specific Perspective-Taking, depicted in Table~\ref{tab:ablation_results} as \textsc{SimToM}-Domain. We find that Domain-Specific perspective taking strategies lead to a substantial performance gain over \textsc{SimToM}, including from 70.5\% to 90.5\% on False Belief tests in BigTOM, a 20\% absolute accuracy improvement (a 34.25\% absolute improvement over 0-shot CoT), and from 72.8 to 79.3 across all ToMI questions, a 6.5\% accuracy improvement. This result suggests that \textbf{LLMs with improved perspective-taking capabilities may display subsantially stronger ToM capabilities}. And although these performances are strong, there is still a gap to ToM based on oracle-level perspective-taking: \textsc{SimToM}-Domain lags behind \textsc{SimToM}-Oracle by over 10\% on ToMI false belief, by 5.5\% on BigTOM false belief, and by over 3 and 6\% on All question types respectively. Future work in ToM may therefore find it fruitful to investigate perspective-taking capabilities more deeply.

\subsection{Extension to Complex Theory-of-Mind Scenarios}
One concern with SimTOM may be that it fails to generalize to more complex, real-world ToM scenarios than the synthetic BigTOM and ToMI datasets. To evaluate this, we investigate our method qualitatively on a small, hand-selected subset of 30 questions drawn from the real-world CosmosQA dataset~\citep{huang2019cosmos} that test ToM capabilities.


On BigTOM and ToMI, perspective-taking takes the form of hiding information because all information is known to the omniscient narrator. However, in a real-world context this will not always be the case, and so perspective-taking may involve more 
\textit{inferring} than \textit{hiding} information.  We observed this to be the case on the CosmosQA dataset. For example, given a question drawn from a blog post about why the speaker's romantic interest did not return a message, GPT-3.5-Turbo outputted the following as the person of interest's perspective: ``it's possible that I wasn't yet completely invested in the relationship or had other priorities such as schoolwork''. Interestingly, in this case the perspective-taking actually infers the key information necessary to answer the question correctly (that the person may have been busy with schoolwork) based on only a tangential mention of studying in the original question. However, this was not always the case; oftentimes model imputations were more ``hallucination'' than ``inference''. In some cases the model took liberties with the framing of the perspective that led to incorrect inferences; in another case the model even outputted a detailed perspective based on an entirely made up premise absent from the question. The relationship between hallucination and inference will be of particular importance when applying perspective-taking to scenarios with limited information, such as many real-world ToM applications.

\section{Related Work}
\paragraph{Evaluating LLM Capabilities on ToM}
Recent approaches on evaluating zero-shot LLM capabilities on ToM have made use of LM probing~\citep{sapNeuralTheoryofMindLimits2022}, often replaced by MC ~\citep{sapNeuralTheoryofMindLimits2022,shapira2023clever} and CoT probing~\citep{wei2022chain,kojima2022large} for instruction-tuned models. Though these prompt-styles can elicit different performance capabilities~\citep{wei2022chain}, we modify our prompts only minimally to evaluate our hypotheses, motivated by ~\citet{shapira2023clever}'s argument that
excessive prompt-tuning violates a ``reasonable expectation'' of evaluation standards for LLMs on ToM.

\paragraph{Enhancing LLM Reasoning Capabilities}
Motivated in part by early results that transformers can engage in reasoning tasks~\citep{clark2020transformers}, a number of prompt-based methods for enhancing LLM reasoning capabilities have been published in recent years, such as ProofWriter~\citep{tafjord2020proofwriter}. ~\citet{shi2023large} point out that LLMs can be easily distracted by irrelevant information during reasoning, and ~\citet{creswell2022selection} use LLMs to select relevant facts before inferring new ones. This selection is somewhat similar to our perspective-taking, but a key distinction lies in that perspective-taking attempts not only to isolate \textit{relevant} facts, but ones that agents \textit{should not know about} when being simulated – it is not clear how to implement \textit{relevance} for selection in ToM.


Recent approaches have also combined LLM reasoning with symbolic reasoning algorithms. ~\citet{nye2021improving} separates ``System 1'' and ``System 2'' into neural- and symbolic-systems respectively to perform reasoning tasks. A relevant contemporaneous work,~\citet{sclar-etal-2023-minding}, uses LLMs to construct a symbolic belief state graph and answers questions using an inference-time graph algorithm. It is not clear, however, how to apply these symbolic methods to naturalistic ToM datasets in such as BigTOM. More broadly, these approaches both represent a distinct direction to ours, as our approach studies how to enable ToM capabilities in LLMs without external memory or human-written inference-time algorithms. 





\section{Conclusion}
In this paper, we propose a two-prompt approach to eliciting Theory-of-Mind capabilities in LLMs by first asking LLMs to \textbf{perspective-take}: to filter context to what the character in question knows. We find that LLMs are substantially stronger at ToM reasoning than they appear when probed with a single inference pass.
Our key takeaways from rigorously analyzing \textsc{SimToM} are that (1) that perspective-taking is best implemented as a separate prompt before question-answering, instead of in a single prompt (Section~\ref{subsec:anal_sim_oneprompt}); 
(2) that \textit{today's} LLMs could close-to-solve ToM benchmarks with access to oracle perspective-taking capabilities (Section~\ref{subsec:anal_oracle});
and (3) that LLMs with improved perspective-taking capabilities could achieve even stronger performance on ToM tasks (Section~\ref{subsec:anal_domain_specific}).

\section{Future Work}
Future work on building ToM capabilities into LLMs may find it fruitful to explore both evaluating and enhancing LLMs' perspective-taking capabilities. 
The idea of perspective-taking has a broader scope in cognitive science~\citep{barlassinaFolkPsychologyMental2017, gordonFolkPsychologySimulation1986} than is currently being tested in simple Sally-Anne false-belief tests, where belief states primarily concern object locations. Further evaluating models' perspective-taking capabilities in more complex, lifelike ToM settings may uncover new challenges in modeling perspective-taking.



\section*{Limitations}
One limitation of our work is that perspective-taking is currently implemented by ``hiding'' parts of the original story when answering the question from an agent's perspective. Perspective-taking may not always involve ``hiding'' information from a complete knowledge of the world state, and in some cases could require feasibly imputing unseen information (e.g., ``Imagine you were in their shoes...''). This is mainly due to a limitation in the datasets we use to test our method; to the best of our knowledge current reading comprehension datasets for ToM are of the Sally-Anne belief form, which tests basic abilities to anticipate beliefs and actions given a change to the world state that we (omniscient viewers) are aware of, but the character is not. Another limitation is that we have not performed tests with smaller language models than 7B parameters; it may be the case that only LLMs are capable of performing ToM to this level using \textsc{SimToM}. As LLMs are improving in performance, however, this may not be a limitation that is particularly disruptive to future research in ToM.

\section*{Acknowledgements}
This material is based upon work partially supported by National Science Foundation awards 1722822 and 1750439, and National Institutes of Health awards R01MH125740, R01MH132225, R01MH096951 and R21MH130767. Any opinions, findings, conclusions, or recommendations expressed in this material are those of the author(s) and do not necessarily reflect the views of the sponsors, and no official endorsement should be inferred. We would also like to thank Daniel Fried for insightful comments and suggestions related to this project.

\section*{Ethics}
Advanced ToM capabilities applied to real world settings could have unethical use cases and modes of operation. Unethical use cases could include inferring mental states without consent, particularly as part of deceptive politically or financially motivated automated persuasion techniques. Unethical modes of operation could include inaccurately, harmfully, or unfairly interacting with users based on ToM inferences in ways that could amplify the negative effects of biases in LLMs. However, there are many upsides of advanced ToM capabilities as well, including the benefits of more empathetic and socially aware technology. As AI seems to be trending towards inclusion in a broader swath of interactive products, advanced ToM could become increasingly impactful to the way we interact with the software systems around us. We would recommend that future research that focuses on use cases that directly interact with people (instead of offline prediction, as in our experiments), particularly in domains or tasks that may be harmful (such as unconscious belief change) carefully consider the distribution of outcomes improved ToM capabilities in that domain would effect.

\bibliography{anthology,custom}

\ifdefined\DeclarePrefChars\DeclarePrefChars{'’-}\else\fi
\begin{thebibliography}{43}
\expandafter\ifx\csname natexlab\endcsname\relax\def\natexlab#1{#1}\fi

\bibitem[{Arodi and Cheung(2021)}]{arodiTextualTimeTravel2021}
Akshatha Arodi and Jackie Chi~Kit Cheung. 2021.
\newblock Textual time travel: A temporally informed approach to theory of
  mind.
\newblock In \emph{Findings of the Association for Computational Linguistics:
  EMNLP 2021}, pages 4162--4172.

\bibitem[{Barlassina and Gordon(2017)}]{barlassinaFolkPsychologyMental2017}
Luca Barlassina and Robert~M. Gordon. 2017.
\newblock {Folk Psychology as Mental Simulation}.
\newblock In Edward~N. Zalta, editor, \emph{The {Stanford} Encyclopedia of
  Philosophy}, {S}ummer 2017 edition. Metaphysics Research Lab, Stanford
  University.

\bibitem[{Baron-Cohen et~al.(1985)Baron-Cohen, Leslie, and
  Frith}]{BARONCOHEN198537}
Simon Baron-Cohen, Alan~M Leslie, and Uta Frith. 1985.
\newblock Does the autistic child have a “theory of mind”?
\newblock \emph{Cognition}, 21(1):37--46.

\bibitem[{Bubeck et~al.(2023)Bubeck, Chandrasekaran, Eldan, Gehrke, Horvitz,
  Kamar, Lee, Lee, Li, Lundberg et~al.}]{bubeck2023sparks}
S{\'e}bastien Bubeck, Varun Chandrasekaran, Ronen Eldan, Johannes Gehrke, Eric
  Horvitz, Ece Kamar, Peter Lee, Yin~Tat Lee, Yuanzhi Li, Scott Lundberg,
  et~al. 2023.
\newblock Sparks of artificial general intelligence: Early experiments with
  gpt-4.
\newblock \emph{arXiv preprint arXiv:2303.12712}.

\bibitem[{Carruthers(2009)}]{carruthers_2009}
Peter Carruthers. 2009.
\newblock \href {https://doi.org/10.1017/S0140525X09000545} {How we know our
  own minds: The relationship between mindreading and metacognition}.
\newblock \emph{Behavioral and Brain Sciences}, 32(2):121–138.

\bibitem[{Clark et~al.(2020)Clark, Tafjord, and
  Richardson}]{clark2020transformers}
Peter Clark, Oyvind Tafjord, and Kyle Richardson. 2020.
\newblock Transformers as soft reasoners over language.
\newblock \emph{arXiv preprint arXiv:2002.05867}.

\bibitem[{Craver(2007)}]{craver2007}
Carl~F. Craver. 2007.
\newblock \emph{Explaining the Brain: Mechanisms and the Mosaic Unity of
  Neuroscience}.
\newblock New York : Oxford University Press,: Oxford University Press,
  Clarendon Press.

\bibitem[{Creswell et~al.(2022)Creswell, Shanahan, and
  Higgins}]{creswell2022selection}
Antonia Creswell, Murray Shanahan, and Irina Higgins. 2022.
\newblock Selection-inference: Exploiting large language models for
  interpretable logical reasoning.
\newblock \emph{arXiv preprint arXiv:2205.09712}.

\bibitem[{Currie(2002{\natexlab{a}})}]{currieRecreativeMindsImagination2002}
Gregory Currie. 2002{\natexlab{a}}.
\newblock \href {https://doi.org/10.1093/acprof:oso/9780198238089.003.0001}
  {{1Introduction}}.
\newblock In \emph{{Recreative Minds}}. Oxford University Press.

\bibitem[{Currie(2002{\natexlab{b}})}]{currieDesireImagination2002}
Gregory Currie. 2002{\natexlab{b}}.
\newblock Desire in imagination.
\newblock In Tamar~Szabo Gendler and John Hawthorne, editors,
  \emph{Conceivability and Possibility}, pages 201--221. Oxford University
  Press.

\bibitem[{Evans(1982)}]{evansVarietiesReference1982}
Gareth Evans. 1982.
\newblock \emph{The Varieties of Reference}.
\newblock Oxford: Oxford University Press.

\bibitem[{Gallese and Goldman(1998)}]{galleseMirrorNeuronsSimulation1998}
Vittorio Gallese and Alvin Goldman. 1998.
\newblock Mirror neurons and the simulation theory of mind-reading.
\newblock \emph{Trends in cognitive sciences}, 2(12):493--501.

\bibitem[{Gallese et~al.(2004)Gallese, Keysers, and
  Rizzolatti}]{galleseUnifyingViewBasis2004}
Vittorio Gallese, Christian Keysers, and Giacomo Rizzolatti. 2004.
\newblock A unifying view of the basis of social cognition.
\newblock \emph{Trends in cognitive sciences}, 8(9):396--403.

\bibitem[{Gandhi et~al.(2023)Gandhi, Fr{\"a}nken, Gerstenberg, and
  Goodman}]{gandhiUnderstandingSocialReasoning2023}
Kanishk Gandhi, Jan-Philipp Fr{\"a}nken, Tobias Gerstenberg, and Noah~D
  Goodman. 2023.
\newblock Understanding social reasoning in language models with language
  models.
\newblock \emph{arXiv preprint arXiv:2306.15448}.

\bibitem[{Goldman(2006)}]{goldman2006}
Alvin~I. Goldman. 2006.
\newblock \emph{Simulating Minds: The Philosophy, Psychology, and Neuroscience
  of Mindreading}.
\newblock New York, US: Oxford University Press USA.

\bibitem[{Goldman(2008)}]{goldman2008}
Alvin~I. Goldman. 2008.
\newblock \href
  {https://doi.org/https://doi.org/10.1111/j.1933-1592.2008.00221.x} {Hurley on
  simulation}.
\newblock \emph{Philosophy and Phenomenological Research}, 77(3):775--788.

\bibitem[{Gopnik and Wellman(1994)}]{gopnikTheoryTheory1994}
Alison Gopnik and Henry~M Wellman. 1994.
\newblock The theory theory.
\newblock In \emph{An earlier version of this chapter was presented at the
  Society for Research in Child Development Meeting, 1991.} Cambridge
  University Press.

\bibitem[{Gordon(1986)}]{gordonFolkPsychologySimulation1986}
Robert~M. Gordon. 1986.
\newblock \href {https://doi.org/10.1111/j.1468-0017.1986.tb00324.x} {Folk
  psychology as simulation}.
\newblock \emph{Mind and Language}, 1(2):158--71.

\bibitem[{Gordon(1995)}]{gordonSimulationIntrospectionInference1995}
Robert~M. Gordon. 1995.
\newblock Simulation without introspection or inference from me to you.
\newblock In Martin Davies and Tony Stone, editors, \emph{Mental Simulation}.
  Blackwell.

\bibitem[{Gordon(2007)}]{gordonAscentRoutinesPropositional2007}
Robert~M. Gordon. 2007.
\newblock \href {https://doi.org/10.1007/s11229-007-9202-9} {Ascent routines
  for propositional attitudes}.
\newblock \emph{Synthese}, 159(2):151--165.

\bibitem[{Heal(1994)}]{heal1994simulation}
Jane Heal. 1994.
\newblock Simulation vs. theory-theory: What is at issue?
\newblock In Christopher Peacocke, editor, \emph{Objectivity, Simulation and
  the Unity of Consciousness: Current Issues in the Philosophy of Mind}. Oxford
  University Press.

\bibitem[{Huang et~al.(2019)Huang, Bras, Bhagavatula, and
  Choi}]{huang2019cosmos}
Lifu Huang, Ronan~Le Bras, Chandra Bhagavatula, and Yejin Choi. 2019.
\newblock Cosmos qa: Machine reading comprehension with contextual commonsense
  reasoning.
\newblock \emph{arXiv preprint arXiv:1909.00277}.

\bibitem[{Hurley(2008)}]{hurley2008sim}
Susan Hurley. 2008.
\newblock \href
  {https://doi.org/https://doi.org/10.1111/j.1933-1592.2008.00220.x}
  {Understanding simulation1}.
\newblock \emph{Philosophy and Phenomenological Research}, 77(3):755--774.

\bibitem[{Kojima et~al.(2022)Kojima, Gu, Reid, Matsuo, and
  Iwasawa}]{kojima2022large}
Takeshi Kojima, Shixiang~Shane Gu, Machel Reid, Yutaka Matsuo, and Yusuke
  Iwasawa. 2022.
\newblock Large language models are zero-shot reasoners.
\newblock \emph{Advances in neural information processing systems},
  35:22199--22213.

\bibitem[{Kosinski(2023)}]{kosinski2023theory}
Michal Kosinski. 2023.
\newblock Theory of mind may have spontaneously emerged in large language
  models.
\newblock \emph{arXiv preprint arXiv:2302.02083}.

\bibitem[{Langley et~al.(2022)Langley, Cirstea, Cuzzolin, and
  Sahakian}]{langley2022theory}
Christelle Langley, Bogdan~Ionut Cirstea, Fabio Cuzzolin, and Barbara~J
  Sahakian. 2022.
\newblock Theory of mind and preference learning at the interface of cognitive
  science, neuroscience, and ai: A review.
\newblock \emph{Frontiers in Artificial Intelligence}, 5:62.

\bibitem[{Le et~al.(2019)Le, Boureau, and Nickel}]{le2019revisiting}
Matthew Le, Y-Lan Boureau, and Maximilian Nickel. 2019.
\newblock Revisiting the evaluation of theory of mind through question
  answering.
\newblock In \emph{Proceedings of the 2019 Conference on Empirical Methods in
  Natural Language Processing and the 9th International Joint Conference on
  Natural Language Processing (EMNLP-IJCNLP)}, pages 5872--5877.

\bibitem[{Nye et~al.(2021)Nye, Tessler, Tenenbaum, and Lake}]{nye2021improving}
Maxwell Nye, Michael Tessler, Josh Tenenbaum, and Brenden~M Lake. 2021.
\newblock Improving coherence and consistency in neural sequence models with
  dual-system, neuro-symbolic reasoning.
\newblock \emph{Advances in Neural Information Processing Systems},
  34:25192--25204.

\bibitem[{OpenAI(2023)}]{openai2023gpt4}
OpenAI. 2023.
\newblock \href {http://arxiv.org/abs/2303.08774} {Gpt-4 technical report}.

\bibitem[{Premack and Woodruff(1978)}]{premack1978does}
David Premack and Guy Woodruff. 1978.
\newblock Does the chimpanzee have a theory of mind?
\newblock \emph{Behavioral and brain sciences}, 1(4):515--526.

\bibitem[{Sap et~al.(2022)Sap, LeBras, Fried, and
  Choi}]{sapNeuralTheoryofMindLimits2022}
Maarten Sap, Ronan LeBras, Daniel Fried, and Yejin Choi. 2022.
\newblock Neural theory-of-mind? on the limits of social intelligence in large
  lms.
\newblock \emph{arXiv preprint arXiv:2210.13312}.

\bibitem[{Sap et~al.(2019)Sap, Rashkin, Chen, Le~Bras, and
  Choi}]{sap-etal-2019-social}
Maarten Sap, Hannah Rashkin, Derek Chen, Ronan Le~Bras, and Yejin Choi. 2019.
\newblock \href {https://doi.org/10.18653/v1/D19-1454} {Social {IQ}a:
  Commonsense reasoning about social interactions}.
\newblock In \emph{Proceedings of the 2019 Conference on Empirical Methods in
  Natural Language Processing and the 9th International Joint Conference on
  Natural Language Processing (EMNLP-IJCNLP)}, pages 4463--4473, Hong Kong,
  China. Association for Computational Linguistics.

\bibitem[{Sclar et~al.(2023)Sclar, Kumar, West, Suhr, Choi, and
  Tsvetkov}]{sclar-etal-2023-minding}
Melanie Sclar, Sachin Kumar, Peter West, Alane Suhr, Yejin Choi, and Yulia
  Tsvetkov. 2023.
\newblock \href {https://aclanthology.org/2023.acl-long.780} {Minding language
  models{'} (lack of) theory of mind: A plug-and-play multi-character belief
  tracker}.
\newblock In \emph{Proceedings of the 61st Annual Meeting of the Association
  for Computational Linguistics (Volume 1: Long Papers)}, pages 13960--13980,
  Toronto, Canada. Association for Computational Linguistics.

\bibitem[{Shapira et~al.(2023)Shapira, Levy, Alavi, Zhou, Choi, Goldberg, Sap,
  and Shwartz}]{shapira2023clever}
Natalie Shapira, Mosh Levy, Seyed~Hossein Alavi, Xuhui Zhou, Yejin Choi, Yoav
  Goldberg, Maarten Sap, and Vered Shwartz. 2023.
\newblock Clever hans or neural theory of mind? stress testing social reasoning
  in large language models.
\newblock \emph{arXiv preprint arXiv:2305.14763}.

\bibitem[{Shi et~al.(2023)Shi, Chen, Misra, Scales, Dohan, Chi, Sch{\"a}rli,
  and Zhou}]{shi2023large}
Freda Shi, Xinyun Chen, Kanishka Misra, Nathan Scales, David Dohan, Ed~H Chi,
  Nathanael Sch{\"a}rli, and Denny Zhou. 2023.
\newblock Large language models can be easily distracted by irrelevant context.
\newblock In \emph{International Conference on Machine Learning}, pages
  31210--31227. PMLR.

\bibitem[{Tafjord et~al.(2020)Tafjord, Mishra, and
  Clark}]{tafjord2020proofwriter}
Oyvind Tafjord, Bhavana~Dalvi Mishra, and Peter Clark. 2020.
\newblock Proofwriter: Generating implications, proofs, and abductive
  statements over natural language.
\newblock \emph{arXiv preprint arXiv:2012.13048}.

\bibitem[{Touvron et~al.(2023)Touvron, Martin, Stone, Albert, Almahairi,
  Babaei, Bashlykov, Batra, Bhargava, Bhosale et~al.}]{touvron2023llama}
Hugo Touvron, Louis Martin, Kevin Stone, Peter Albert, Amjad Almahairi, Yasmine
  Babaei, Nikolay Bashlykov, Soumya Batra, Prajjwal Bhargava, Shruti Bhosale,
  et~al. 2023.
\newblock Llama 2: Open foundation and fine-tuned chat models.
\newblock \emph{arXiv preprint arXiv:2307.09288}.

\bibitem[{Ullman(2023)}]{ullman2023large}
Tomer Ullman. 2023.
\newblock Large language models fail on trivial alterations to theory-of-mind
  tasks.
\newblock \emph{arXiv preprint arXiv:2302.08399}.

\bibitem[{Wang et~al.(2022)Wang, Wei, Schuurmans, Le, Chi, Narang, Chowdhery,
  and Zhou}]{wang2022self}
Xuezhi Wang, Jason Wei, Dale Schuurmans, Quoc Le, Ed~Chi, Sharan Narang,
  Aakanksha Chowdhery, and Denny Zhou. 2022.
\newblock Self-consistency improves chain of thought reasoning in language
  models.
\newblock \emph{arXiv preprint arXiv:2203.11171}.

\bibitem[{Wei et~al.(2022)Wei, Wang, Schuurmans, Bosma, Xia, Chi, Le, Zhou
  et~al.}]{wei2022chain}
Jason Wei, Xuezhi Wang, Dale Schuurmans, Maarten Bosma, Fei Xia, Ed~Chi, Quoc~V
  Le, Denny Zhou, et~al. 2022.
\newblock Chain-of-thought prompting elicits reasoning in large language
  models.
\newblock \emph{Advances in Neural Information Processing Systems},
  35:24824--24837.

\bibitem[{Wellman et~al.(2001)Wellman, Cross, and Watson}]{wellman2001}
Henry~M. Wellman, David Cross, and Julanne Watson. 2001.
\newblock \href {https://doi.org/https://doi.org/10.1111/1467-8624.00304}
  {Meta-analysis of theory-of-mind development: The truth about false belief}.
\newblock \emph{Child Development}, 72(3):655--684.

\bibitem[{Yao et~al.(2023)Yao, Yu, Zhao, Shafran, Griffiths, Cao, and
  Narasimhan}]{yao2023tree}
Shunyu Yao, Dian Yu, Jeffrey Zhao, Izhak Shafran, Thomas~L Griffiths, Yuan Cao,
  and Karthik Narasimhan. 2023.
\newblock Tree of thoughts: Deliberate problem solving with large language
  models.
\newblock \emph{arXiv preprint arXiv:2305.10601}.

\bibitem[{Zhao et~al.(2023)Zhao, Zhou, Li, Tang, Wang, Hou, Min, Zhang, Zhang,
  Dong et~al.}]{zhao2023survey}
Wayne~Xin Zhao, Kun Zhou, Junyi Li, Tianyi Tang, Xiaolei Wang, Yupeng Hou,
  Yingqian Min, Beichen Zhang, Junjie Zhang, Zican Dong, et~al. 2023.
\newblock A survey of large language models.
\newblock \emph{arXiv preprint arXiv:2303.18223}.

\end{thebibliography}
\bibliographystyle{acl_natbib}

\appendix
\section{A Further Discussion of Cognitive Science Perspectives on Simulation Theory}
\label{sec:appendix_st_tt}
ST is both philosophically and empirically supported, presenting a viable theory for ToM capabilities in humans. Philosophical support for ST comes from ~\citep{gordonAscentRoutinesPropositional2007}, who argue that the process of reporting others' beliefs can be achieved by reporting what is observed from their perspective. This relies on work by ~\citep{evansVarietiesReference1982}, who argue that the process of performing ToM on oneself is not introspective but is achieved by observing external circumstances – e.g., to answer the question ``Do I believe that $p$'', the question asked is, "is it the case that $p$?". Consequently, to perform ToM on others, one needs to simulate being them, looking externally from their perspective~\citep{gordonFolkPsychologySimulation1986, gordonSimulationIntrospectionInference1995}.  There is also evidence from the neuroscience community, particularly from the discovery of mirror neurons, that simulation theory may be be grounded in a measurable physical phenomenon~\citep{galleseMirrorNeuronsSimulation1998,galleseUnifyingViewBasis2004, hurley2008sim}. 

According to~\citet{heal1994simulation}, the implementation of these cognitive mechanisms is irrelevant as long as it results in similar mental states. This suggests that ST capabilities may be possible to achieve using \textit{artificial} cognitive processes such as those found in LLMs, motivating our approach.

ST's main alternative within the ToM literature is ``Theory-Theory'' (TT)~\citep{gopnikTheoryTheory1994}: the idea that humans perform ToM by inferring from a large body of commonsense rules and knowledge for how the world works: e.g. the ``Law of Sight'': if a person has no visual impairments or occlusions and something is in front of them, they will see it ~\citep{barlassinaFolkPsychologyMental2017}. The debate between ST and TT has raged for decades; in this work, we do not weigh in on whether one works over another. Instead, we implement to the best of our abilities an explicitly simulation-based ToM approach and report our results on well established ToM tasks.

\section{\textsc{SimToM} Prompts}
\label{sec:appendix_prompts}
Our \textsc{SimToM} prompts are reproduced here in their entirety. Our prompts slightly differ based on model architecture due to the need to accommodate to the differing nature and format of the model outputs. 

For example, Llama-2's safety guardrails often result in the model saying that there is not enough information in the question, and thus cannot respond with a non-factual answer. Hence, we add lines such as "Do not say there is not enough information. Answer with a single word, do not output
anything else" in order to mitigate this phenomenon. However, we maintain that the core content and instructions of the prompts are essentially the same.

\subsection{ToMI Prompts}
\label{sec:appendix_tomi_prompts}

\subsubsection{Perspective Taking}
\begin{llmprompt}[GPT/Llama-2-chat]
The following is a sequence of events about some characters, that takes place in multiple locations.

Your job is to output only the events that the specified character, \{character\}, knows about.

Here are a few rules:

1. A character knows about all events that they do.

2. If a character is in a certain room/location, that character knows about all other events that happens in the room. This includes other characters leaving or exiting the location, the locations of objects in that location, and whether somebody moves an object to another place.

3. If a character leaves a location, and is NOT in that location, they no longer know about any events that happen within that location. However, they can re-enter the location.
\smallskip

Story:
\{story\}
\smallskip

What events does \{character\} know about? Only output the events according to the above rules, do not provide an explanation.
\end{llmprompt}

\subsubsection{Simulation}

\begin{llmprompt}[GPT]
\{perspective\}
\smallskip

You are \{name\}.
\smallskip

Based on the above information, answer the following question:
\smallskip

\{question\}
\smallskip

Keep your answer concise, one sentence is enough. You must choose one of the above choices.
\end{llmprompt}

\begin{llmprompt}[Llama-2-chat]
\{perspective\}
\smallskip

You are \{name\}.
\smallskip

Based on the above information, answer the following question:
\smallskip

\{question\}
\smallskip

You must choose one of the above choices, do not say there is not enough information. Answer with a single word, do not output anything else. 
\end{llmprompt}

\subsection{BigTOM Prompts}
\label{sec:appendix_bigtom_prompts}

\subsubsection{Perspective Taking}
\begin{llmprompt}[GPT]
Imagine you are \{name\}, and consider this story that has an unexpected event. \smallskip

\{story\} \smallskip

If the last sentence of the story says \{name\} notices, sees or realizes the unexpected event in this story, simply output the original story with nothing changed. \smallskip

However, if the sentence says you are not aware of the changes in this story, output only the events you know, i.e., the sentences before the unexpected event happens. \smallskip

Output either the original story or the edited story, nothing else. \smallskip

Format your answer as follows: \smallskip

Sees/Notices/Realizes: (Yes/No) \smallskip

Story:
\end{llmprompt}

\begin{llmprompt}[Llama-2-chat]
Consider this story with an unexpected event.\smallskip

\{story\}\smallskip

Does the story say that \{name\} notices/sees/realizes the unexpected event?\smallskip

If so, simply output the original story with nothing changed. 
However, if \{name\} is not aware of the changes in this story, output only the events that \{name\} knows, i.e., the events before the unexpected event happens.
\end{llmprompt}

\subsubsection{Simulation}

\begin{llmprompt}[GPT]
\{perspective\} \smallskip

You are \{name\}. \smallskip

Based on the above information, answer the following question: \smallskip

\{question\} \smallskip

Answer the questions based on the context. Keep your answer concise, few words are enough, maximum one sentence. Answer as 'Answer:<option>)<answer>'.
\end{llmprompt}

\begin{llmprompt}[Llama-2-chat]
Answer the questions based on the context. Keep your answer concise, few words are enough, maximum one sentence. Answer as 'Answer:<option>)<answer>'. \smallskip

\{perspective\} \smallskip

You are \{name\}. \smallskip

\{question\} \smallskip

Choose the most straightforward answer.
\end{llmprompt}

\section{Single Prompts}
\label{sec:appendix_single_prompts}
\begin{llmprompt}[ToMI]
I will give you a sequence of events about some characters, that takes place in multiple locations. and a question that asks about the sequence of events. Your task is in two steps. \\
Step 1. You will first output only the events that the specified character, \{character\}, knows about. \\
Here are a few rules: \\
    1. A character knows about all events that they do. \\
    2. If a character is in a certain room/location, that character knows about all other events that happens in the room. This includes other characters leaving or exiting the location, the locations of objects in that location, and whether somebody moves an object to another place. \\
    3. If a character leaves a location, and is NOT in that location, they no longer know about any events that happen within that location. However, they can re-enter the location. \\
Step 2. You will then imagine you are the main character, \{character\}, then answer the question given to you based on the story you have rewritten. Ignore the previous sequence of events -- your rewritten sequence of events are now the new events. Do not output a blank answer or say you do not have enough information. \\
Do Step 1 and Step 2 combined. \\
Story: \{story\} \\
Output the sentences that only \{character\} knows about. \\
Question: \{question\} \\
Format your answer as follows: \\
Step 1: (list of events) \\
Step 2: Answer: (answer to question) \\
\end{llmprompt}

\begin{llmprompt}[BigTOM]
I will give you a short story, and a question that asks about the story. Your task is in two steps. \\
Step 1: Imagine you are \{name\}, and consider this story that has an unexpected event. \\
\{story\} \\
If the last sentence of the story says \{name\} notices, sees or realizes the unexpected event in this story, simply output the original story with nothing changed. \\
However, if the sentence says you are not aware of the changes in this story, output only the events you know, i.e., the sentences before the unexpected event happens. \\
Output either the original story or the edited story, nothing else. \\
Format your answer for step 1 as follows: \\
Sees/Notices/Realizes: (Yes/No) \\
Story: \\
2. You will then imagine you are the main character, \{name\}, then answer the question given to you based on the story you have rewritten. Ignore the previous story -- your rewritten story is now the new story. Do not output a blank answer or say you do not have enough information -- you must choose either choice a) or choice b). Answer as  '<option>) <answer>'. \\
Here is the story and question. \\
Story: \{story\} \\
Question: \{question\} \\
Format your answer as follows: \\
Step 1: \\
Step 2:  \\
<answer> \\
\end{llmprompt}

\section{Domain Specific (Few-Shot) Prompts}
\label{sec:appendix_domain_specific_prompts}
\begin{llmprompt}[ToMI]
The following is a sequence of events about some characters, that takes place in multiple locations. \\
A character does not know about any events before they enter a location. \\
If a character is in a certain location, the character knows about the location of all objects within that location. The character also knows if other agents enter or leave that location. They also know if other agents move objects around. \\
If a character leaves that location, they no longer know about anything that occurs within that location, or changes in locations of objects. \\
However, note that a character can re-enter a location. \\
Here are a few examples. \\
Story: \\
1 Lily entered the dining room. \\
2 William entered the dining room. \\
3 The underpants is in the box. \\
4 The box is in the dining room. \\
5 William exited the dining room. \\
6 Abigail entered the cellar. \\
7 William dislikes the eggplant \\
8 Abigail exited the cellar. \\
9 William entered the dining room. \\
10 Lily moved the underpants to the suitcase. \\
11 The suitcase is in the dining room. \\
What events does William know about? \\
William knows about the following events: \\
2 William entered the dining room. \\
3 The underpants is in the box. \\
4 The box is in the dining room. \\
5 William exited the dining room. \\
9 William entered the dining room. \\
10 Lily moved the underpants to the suitcase. \\
11 The suitcase is in the dining room. \\
... (3x) \\
Story: \\
\{story\} \\
What events does \{character\} know about? \\
\end{llmprompt}

\begin{llmprompt}[BigTOM]
I will give you an excerpt. Your task is three steps: \\
1. There is a sentence that describes how the situation unexpectedly changed. Identify this sentence. \\
2. Identify if the main character comes to know about, or notices, this change at the end. \\
3. If the main character does not know about this change, edit the excerpt and output the part of the excerpt BEFORE the sentence that describes the change. If the main character does know about the change, do not edit the excerpt, and output the original story. \\
Here are some examples. \\
Story: Olumide, a skilled woodcarver in a Nigerian village, is preparing to carve a beautiful sculpture for the village chief. Olumide wants to use a sharp chisel to create intricate details on the sculpture. Olumide observes his set of chisels and sees one that appears to be sharp and in perfect condition. However, while Olumide is talking to a fellow artisan, a child from the village accidentally drops the chisel, causing it to become blunt and damaged. Olumide does not notice the damaged chisel on the ground. \\
Sentence: However, while Olumide is talking to a fellow artisan, a child from the village accidentally drops the chisel, causing it to become blunt and damaged. \\
Knows about or notices change: No \\
Edit: Olumide, a skilled woodcarver in a Nigerian village, is preparing to carve a beautiful sculpture for the village chief. Olumide wants to use a sharp chisel to create intricate details on the sculpture. Olumide observes his set of chisels and sees one that appears to be sharp and in perfect condition. \\
Story: \\
\{story\} \\
\end{llmprompt}



\section{Baseline Prompts}
\label{sec:appendix_baseline_prompts}
\begin{llmprompt}[0-shot]
Answer the questions based on the context. Keep your answer concise, few words are enough, maximum one sentence. Answer as 'Answer:<option>)<answer>'.
\end{llmprompt}

\begin{llmprompt}[0-shot CoT]
Answer the questions based on the context. Reason step by step before answering in 'Thought: Let's think step by step'. Write your final answer as 'Answer:<option>)<answer>'. Always pick an option, do not say none of the above or that there is not enough information. \\
\{question\} \\
\{answer choices\} \\
\end{llmprompt}

\section{Non-Instruction Tuned Models}
\label{sec:appendix_non_instruction_tuned_models}
Non-instruction tuned models, such as Llama-2-7b or Llama-2-13b, cannot be prompted with the same instructional prompts. These models are usually evaluated by 1) setting up a prompt such that a natural continuation of the prompt will illicit the desired response, or 2) giving several few-shot examples in the desired format.

However, in investigations done with Llama-2-13b and Llama-2-7b, the former method resulted not in perspective taking, but instead a continuation of the question that the prompt poses, while the latter method would no longer be a zero-shot method of evaluating theory of mind.

Thus, we limit our investigation to fine-tuned instruction-following models.








\section{Breakdown of Results Across Question Types}
\label{sec:appendix_breakdown}
\begin{table*}[h!]

\label{tab:full_results_bigtom}

\centering

\centering
\caption{Results on BigTOM dataset. sim* methods are ablations of \textsc{SimToM}, -dom is the domain-specific perspective-taking, -1prompt is the single prompt ablation. SC-CoT is self-consistency chain of thought~\citep{wang2022self} and ToT is Tree-of-Thoughts~\citep{yao2023tree}.}

\begin{tabular}{llrrrrrrr}
\toprule
          model &            method &    fb &   all &    tb &  action-fb &  action-tb &  belief-fb &  belief-tb \\
\midrule
 Llama2-7b-chat &            0-shot & 47.50 & 53.62 & 59.75 &      41.50 &      68.00 &      53.50 &      51.50 \\
 Llama2-7b-chat &        0-shot CoT & 31.50 & 48.62 & 65.75 &      23.50 &      70.00 &      39.50 &      61.50 \\
 Llama2-7b-chat &            \textsc{SimToM} & 70.50 & 57.25 & 44.00 &      66.00 &      50.50 &      75.00 &      37.50 \\
Llama2-13b-chat &            0-shot & 41.25 & 51.38 & 61.50 &      36.00 &      59.00 &      46.50 &      64.00 \\
Llama2-13b-chat &        0-shot CoT & 52.25 & 56.00 & 59.75 &      52.00 &      57.50 &      52.50 &      62.00 \\
Llama2-13b-chat &            \textsc{SimToM} & 61.75 & 58.00 & 54.25 &      61.00 &      55.50 &      62.50 &      53.00 \\
  gpt-3.5-turbo &            0-shot & 41.00 & 66.38 & 91.75 &      12.50 &      96.00 &      69.50 &      87.50 \\
  gpt-3.5-turbo &        0-shot CoT & 56.25 & 75.88 & 95.50 &      41.00 &      96.00 &      71.50 &      95.00 \\
  gpt-3.5-turbo &            \textsc{SimToM} & 70.50 & 81.62 & 92.75 &      63.00 &      95.50 &      78.00 &      90.00 \\
  gpt-3.5-turbo & sim-1prompt & 50.75 & 54.75 & 58.75 &      46.00 &      69.50 &      55.50 &      48.00 \\
gpt-3.5-turbo & SC-CoT & 54.75 & 75.63 & 96.50 & 30.50 & 98.00 & 79.00 & 95.00 \\
gpt-3.5-turbo & ToT & 15.75 & 53.88 & 92.00 & 8.50 & 94.50 & 23.00 & 89.50 \\

  gpt-3.5-turbo &           sim-dom & 90.50 & 91.50 & 92.50 &      86.00 &      91.00 &      95.00 &      94.00 \\
          gpt-4 &            0-shot & 89.00 & 92.50 & 96.00 &      79.00 &      96.00 &      99.00 &      96.00 \\
          gpt-4 &        0-shot CoT & 93.25 & 95.50 & 97.75 &      87.50 &      98.00 &      99.00 &      97.50 \\
          gpt-4 &            \textsc{SimToM} & 92.00 & 95.00 & 98.00 &      90.00 &      98.00 &      94.00 &      98.00 \\
          gpt-4 &           sim-dom & 96.25 & 95.38 & 94.50 &      98.00 &      95.00 &      94.50 &      94.00 \\
\bottomrule
\end{tabular}

\end{table*}

\begin{table*}[h!]

\label{tab:full_results_tomi}

\centering

\centering
\caption{Results on ToMI dataset. sim* methods are ablations of \textsc{SimToM}, -dom is the domain-specific perspective-taking, -1prompt is the single prompt ablation. *-rules is the 0-shot versions with the same rules as added during \textsc{SimToM} prompting to describe ToMI's quirks. We find that this does not perform differently than without adding these rules, verifying that the rules, while helpful for our perspective-taking approach, do not artificially inflate our method's performance relative to the baselines. The columns are the different question types: nt means ``no-tom'', a control question type asking about a person that has not witnessed a change. Mem-real are the average of the memory and reality question type scores.}

\begin{tabular}{llrrrrrrrr}
\toprule
          model &            method &    fb &   all &    tb &  fo-nt &  fo-t &  so-nt &  so-t &  mem-real \\
\midrule
 Llama2-7b-chat &            0-shot & 28.25 & 44.50 & 50.75 &  49.00 & 29.00 &  52.50 & 27.50 &     64.50 \\
 Llama2-7b-chat &        0-shot CoT & 24.00 & 43.70 & 58.75 &  66.50 & 23.50 &  51.00 & 24.50 &     53.00 \\
 Llama2-7b-chat &            \textsc{SimToM} & 40.00 & 48.10 & 46.50 &  49.50 & 45.00 &  43.50 & 35.00 &     67.50 \\
Llama2-13b-chat &            0-shot & 39.25 & 51.00 & 50.25 &  66.00 & 43.50 &  34.50 & 35.00 &     76.00 \\
Llama2-13b-chat &        0-shot CoT & 16.50 & 45.00 & 63.50 &  70.50 & 15.50 &  56.50 & 17.50 &     65.00 \\
Llama2-13b-chat &            \textsc{SimToM} & 35.50 & 61.10 & 72.00 &  70.50 & 37.00 &  73.50 & 34.00 &     90.50 \\
  gpt-3.5-turbo &            0-shot & 67.25 & 68.60 & 54.25 &  73.50 & 64.00 &  35.00 & 70.50 &    100.00 \\
  gpt-3.5-turbo &        0-shot CoT & 34.00 & 64.10 & 77.50 &  85.50 & 31.50 &  69.50 & 36.50 &     97.50 \\
  gpt-3.5-turbo &       0-shot-rules & 71.50 & 66.80 & 48.25 &  58.00 & 69.50 &  38.50 & 73.50 &     94.50 \\
  gpt-3.5-turbo &            \textsc{SimToM} & 81.00 & 72.80 & 51.00 &  64.50 & 85.00 &  37.50 & 77.00 &    100.00 \\
  gpt-3.5-turbo &         CoT-rules & 78.75 & 66.60 & 48.00 &  65.50 & 82.00 &  30.50 & 75.50 &     79.50 \\
  gpt-3.5-turbo & sim-1prompt & 58.75 & 67.50 & 60.00 &  66.00 & 71.50 &  54.00 & 46.00 &    100.00 \\
  gpt-3.5-turbo &           sim-dom & 85.50 & 79.30 & 62.75 &  78.50 & 89.00 &  47.00 & 82.00 &    100.00 \\
gpt-3.5-turbo & SC-CoT & 33.50 & 66.30 & 80.50 & 87.00 & 29.00 & 74.00 & 38.00 & 88.50 \\
gpt-3.5-turbo & ToT & 25.75 & 59.20 & 80.00 & 77.00 & 33.00 & 83.00 & 18.50 & 84.50 \\
          gpt-4 &            0-shot & 25.50 & 66.50 & 90.75 &  99.50 &  2.00 &  82.00 & 49.00 &    100.00 \\
          gpt-4 &        0-shot CoT & 74.25 & 74.40 & 61.75 &  84.50 & 63.00 &  39.00 & 85.50 &    100.00 \\
          gpt-4 &            \textsc{SimToM} & 87.75 & 87.80 & 81.75 &  92.50 & 95.00 &  71.00 & 80.50 &    100.00 \\
          gpt-4 &           sim-dom & 91.50 & 90.70 & 85.25 &  96.00 & 98.00 &  74.50 & 85.00 &    100.00 \\
\bottomrule
\end{tabular}

\end{table*}

\end{document}